\documentclass[letterpaper, 10pt]{article}

\usepackage[utf8]{inputenc}
\usepackage[T1]{fontenc}
\usepackage{times}
\usepackage[margin=1.0in, letterpaper]{geometry}
\usepackage{titlesec}
\usepackage{graphicx}
\usepackage[labelfont=bf, font=small, skip=3pt, labelsep=period]{caption}
\usepackage{booktabs}
\usepackage{amsmath}
\usepackage{amsfonts}
\usepackage{amssymb}
\usepackage{float}
\usepackage{placeins}
\usepackage{url}
\usepackage{enumitem}
\usepackage{hyperref}

\titleformat{\section}{\normalfont\large\bfseries}{}{0pt}{}
\titleformat{\subsection}{\normalfont\normalsize\itshape}{}{0pt}{}
\titleformat{\subsubsection}{\normalfont\normalsize\itshape}{}{0pt}{}

\titlespacing*{\section}{0pt}{7pt}{3pt}
\titlespacing*{\subsection}{0pt}{5pt}{2pt}
\titlespacing*{\subsubsection}{0pt}{3pt}{1pt}

\newcommand{\cit}[1]{\textsuperscript{\fontsize{8}{8}\selectfont #1}}

\begin{document}
\pagestyle{empty}

\begin{center}
    {\fontsize{14}{16}\selectfont \textbf{Unmasking Toxic Mimicry in Medical Offline Reinforcement Learning for ICU Sepsis Management via Counterfactual Clinical Audits}}
    
    \vspace{0.6em}
    {\fontsize{12}{14}\selectfont \textbf{Hangqi Ren\textsuperscript{1}, Junyi Liao\textsuperscript{2}}}
    
    \vspace{0.3em}
    {\fontsize{12}{14}\selectfont \textsuperscript{1}School of Engineering, Vanderbilt University, Nashville, TN, USA;\\
    \textsuperscript{2}Pratt School of Engineering, Duke University, Durham, NC, USA}
\end{center}

\vspace{0.2em}

\section*{Abstract}
\textit{Offline reinforcement learning (RL) offers considerable promise for optimizing ICU treatment decisions, yet standard evaluation metrics Mean Squared Error (MSE) and Fitted Q-Evaluation (FQE) assess only behavioral imitation and cannot detect Toxic Mimicry, a failure mode in which agents replicate harmful patterns such as treatment withdrawal during comfort-care transitions. Using the MIMIC-III database, we propose the Counterfactual Clinical Audit (CCA) framework, which stress-tests RL agents through physiological perturbations anchored in Surviving Sepsis Campaign (SSC) guidelines. We audit a Medical Decision Transformer (MedDT) and a Historical Causal Transformer (HCT-RL), the latter employing Causal Action Shielding, propensity-based importance weighting, and Conservative Q-Learning. CCA reveals that MedDT paradoxically reduces vasopressor dosage as lactate escalates, contradicting resuscitation guidelines, while HCT-RL maintains physiologically consistent responses. These findings expose a systemic misalignment between statistical fit and clinical safety, supporting counterfactual audits as a necessary evaluation standard for medical RL.}

\vspace{0.2em}
\textbf{Keywords:} Causal Inference, Deep Learning, Critical Care, Artificial Intelligence, Clinical Decision Support, Evaluation

\section{Introduction}

Offline reinforcement learning (RL) has emerged as a promising approach for clinical decision support in intensive care units (ICUs). By learning treatment policies from historical electronic health records (EHRs), RL agents can optimize sequential decisions to maximize long-term patient outcomes.\cit{1,2} The AI Clinician study demonstrated that RL-derived policies could potentially improve sepsis management,\cit{3} and subsequent work has employed Deep Q-Networks, actor-critic methods, and Decision Transformers with encouraging simulation results.\cit{4--7}

Despite this promise, clinical deployment of medical RL remains far from reality.\cit{8,9} The core barrier lies in evaluation methodology. Current benchmarks rely on predictive accuracy (MSE) or value-based proxies such as Fitted Q-Evaluation (FQE),\cit{10,11} which assess how closely an agent's actions align with historical clinician behavior but are insensitive to the underlying causal logic of interventions.\cit{9,12} High-fidelity imitation can yield favorable scores even when the imitated behaviors stem from non-therapeutic patterns. This limitation reflects a broader challenge in causal inference from observational data, where spurious associations may be misinterpreted as valid clinical strategies.\cit{13,14}

In ICU hemodynamic support, we identify a clinically significant confounder. Sepsis-3 defines septic shock as requiring vasopressors to maintain mean arterial pressure (MAP) $\geq$65 mmHg and a serum lactate level greater than 2 mmol/L, despite adequate fluid resuscitation.\cit{15} The Surviving Sepsis Campaign (SSC) guidelines further recommend lactate-guided resuscitation and vasopressor titration to maintain hemodynamic targets.\cit{16} However, vasopressor withdrawal in real-world practice frequently occurs not from physiological improvement but from transitions to comfort care.\cit{17,18} Studies document that withdrawal of life-sustaining treatment varies from 0\% to 84\% among deceased ICU patients.\cit{19} These conditions combine to create a systematic confound in historical ICU data: guidelines mandate vasopressor escalation, yet palliative withdrawal produces the opposite pattern in observational records, with no goal-of-care labels to distinguish the two. We term this the palliative confounder.

Without causal grounding, high-capacity sequence models risk interpreting care-withdrawal patterns as valid responses to metabolic stress. We define this failure mode as Toxic Mimicry: agents inadvertently learn harmful clinical patterns as optimal strategies, recommending decreased support precisely when guidelines mandate intensification.

To address this safety gap, we propose the Counterfactual Clinical Audit (CCA) framework, which stress-tests policies through controlled physiological perturbations anchored in clinical guidelines. CCA comprises three audits: Spurious Robustness, which tests invariance to non-clinical noise; Causal Trend, which verifies dose-response alignment with guidelines; and Contextual Scissor Probe, which evaluates urgency scaling based on shock severity.

We validate CCA by auditing two architectures using the MIMIC-III database:\cit{20,21} a standard Medical Decision Transformer (MedDT)\cit{6} and a Historical Causal Transformer (HCT-RL) incorporating Causal Action Shielding. Our contributions are: (1) characterizing Toxic Mimicry as a critical failure mode arising from the palliative confounder; (2) introducing CCA as a necessary complement to standard evaluation metrics; and (3) demonstrating that causal shielding and importance weighting effectively suppress palliative confounders where standard architectures fail.

\section{Methods}

\subsection{Dataset and Preprocessing}
We utilized the MIMIC-III Clinical Database (v1.4)\cit{20,21} to extract adult patients meeting Sepsis-3 criteria.\cit{15} Our pipeline follows Komorowski et al.\cit{3} The state space $\mathcal{S} \in \mathbb{R}^{44}$ encompasses vital signs, laboratory results, and demographics. Unlike the original discrete formulation, we defined a continuous action space $\mathcal{A} \in \mathbb{R}^2$ representing intravenous fluid and vasopressor dosages. We used a sliding window of $L=6$ time steps (each 4 hours), with 70\%/15\%/15\% train/validation/test splits at the patient level. All features were z-score normalized.

\begin{figure}[!t]
    \centering
    \includegraphics[width=0.88\textwidth]{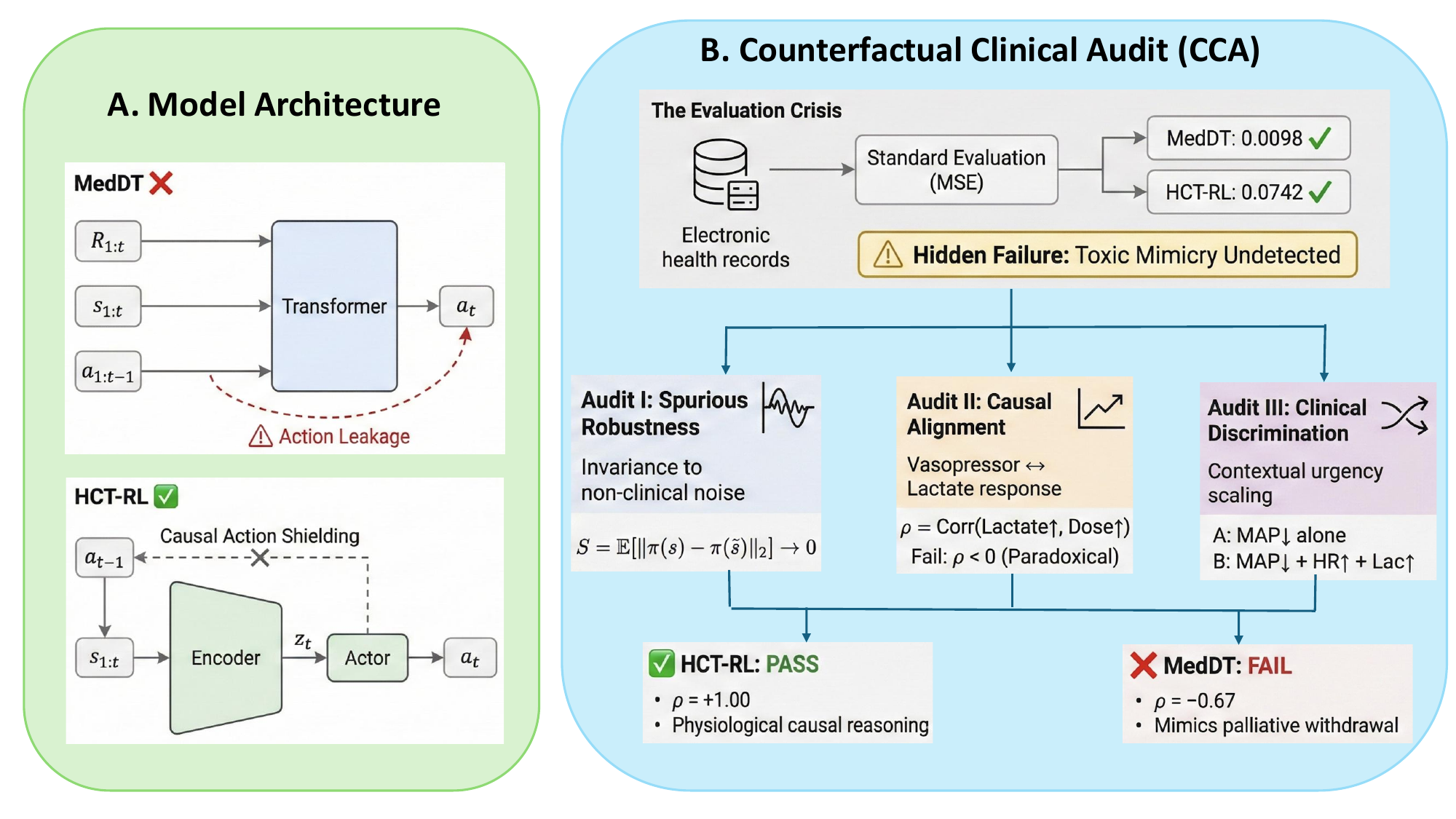}
    \caption{Overview of the proposed framework. (A) Model architectures: MedDT conditions on historical actions $a_{1:t-1}$, creating vulnerability to action leakage. HCT-RL employs Causal Action Shielding to reconstruct policies solely from physiological states. (B) The CCA framework: standard evaluation (MSE) fails to detect Toxic Mimicry. CCA comprises three audits that stress-test policies against clinical principles.}
    \label{fig:framework}
\end{figure}

\subsection{Counterfactual Clinical Audit (CCA) Framework}

Standard metrics (MSE, FQE) assess behavioral imitation but cannot detect adherence to clinical principles. We propose CCA, which stress-tests RL agents through controlled physiological perturbations anchored in SSC guidelines.\cit{16} The framework comprises three complementary audits (Figure 1B).

\subsubsection{Audit I: Spurious Robustness}
Audit I targets susceptibility to spurious correlations, a failure mode in which models respond to non-causal laboratory features that co-vary with historical treatment patterns rather than anchoring decisions to hemodynamically relevant signals. Given a policy $\pi$ and test set $\mathcal{D}_{\text{test}}$, we inject Gaussian noise $\epsilon \sim \mathcal{N}(0, \sigma^2)$ independently at each of the $L$ timesteps of a target feature $k$ (serum creatinine, $\sigma=0.5$) while holding all hemodynamic variables constant. We select a non-critical feature because sensitivity to hemodynamically irrelevant perturbations is the specific failure mode under test; sensitivity to critical features (e.g., MAP) is clinically expected and does not constitute a spurious response. For each trajectory $s_{1:t}$, the perturbed trajectory $\tilde{s}_{1:t}$ has feature $k$ replaced by $s_{i,k} + \epsilon_i$ at every step $i$. The population sensitivity score is:
\begin{equation}
S = \mathbb{E}_{s_{1:t} \sim \mathcal{D}_{\text{test}}} \left[ \| \pi(s_{1:t}) - \pi(\tilde{s}_{1:t}) \|_2 \right]
\end{equation}
A clinically sound agent should exhibit low sensitivity ($S \to 0$) to non-critical features. To facilitate cross-model comparison, we compute a relative robustness score $R_i = 1/(1+S_i)$ and normalize: $\bar{R}_i = R_i / \max_j R_j \times 100\%$, where the model with lower sensitivity attains 100\%.

\subsubsection{Audit II: Causal Trend Alignment}
Audit II directly targets the palliative confounder, testing whether the learned dose-response relationship aligns with SSC guidelines, which recommend lactate-guided resuscitation and vasopressor titration to maintain hemodynamic targets in septic shock.\cit{16} For each test trajectory, we counterfactually replace lactate values at all $L$ timesteps with a fixed level $\ell$, generating $\tilde{s}_{1:t}^{(\ell)}$. This sustained-exposure design simulates a patient maintained at a given metabolic state throughout the observation window. We sweep $\ell$ across the physiological range $\mathcal{L}$ (15 equally-spaced values from Z-score $-2$ to $+5$) while holding all other variables at observed values. For each level, we compute the population-average treatment intensity as the L2 norm of the full action vector:
\begin{equation}
\bar{A}(\ell) = \mathbb{E}_{s_{1:t} \sim \mathcal{D}_{\text{test}}} \left[ \left\| \pi\!\left(\tilde{s}_{1:t}^{(\ell)}\right) \right\|_2 \right]
\end{equation}
Results are expressed as percentage change relative to the baseline response at $\ell = \ell_{\min}$. We then compute the Spearman rank correlation between the 15 lactate levels and their corresponding population-average intensities: $\rho = \text{Corr}(\{\ell_1,\ldots,\ell_{15}\},\, \{\bar{A}(\ell_1),\ldots,\bar{A}(\ell_{15})\})$. A positive correlation ($\rho > 0$) indicates causal alignment with guidelines, whereas a negative correlation ($\rho < 0$) signals Toxic Mimicry.

\subsubsection{Audit III: Contextual Scissor Probe}
Audit III tests for contextual blindness, the failure to adjust treatment urgency when hypotension coincides with multi-organ deterioration. A clinically capable agent should intensify vasopressor dosing more aggressively when hypotension co-occurs with multi-organ stress than when it occurs in isolation. We define two physiological scenarios evaluated across a MAP sweep of 10 equally-spaced steps from Z-score $+1$ to $-2$ (high to critically low blood pressure).

Scenario A (Univariate): only MAP is varied while heart rate (HR) and lactate are held at observed values, representing isolated hypotension without systemic deterioration.

Scenario B (Holistic): MAP, HR, and lactate are co-varied simultaneously. As MAP decreases from Z=$+1$ to Z=$-2$, HR increases from Z=$-0.5$ to Z=$+2$ and lactate increases from Z=$-0.5$ to Z=$+2$, modeling the progressive hemodynamic collapse characteristic of distributive shock.

For each scenario, we record the population-mean vasopressor dosage $\bar{A}_{\text{scenario}}(\text{MAP})$ across the MAP sweep and fit a linear regression to obtain response slopes $\beta_{\text{uni}}$ and $\beta_{\text{hol}}$. Scissor Divergence is defined as:
\begin{equation}
\Delta_{\text{scissor}} = |\beta_{\text{hol}}| - |\beta_{\text{uni}}|
\end{equation}
Here $\beta$ represents the linear rate of vasopressor dose change per unit decrease in MAP; a steeper slope (larger $|\beta|$) indicates more aggressive dose escalation as blood pressure falls. A clinically appropriate agent should exhibit positive Scissor Divergence ($\Delta_{\text{scissor}} > 0$), steepening its vasopressor response when hypotension is accompanied by tachycardia and hyperlactatemia compared to isolated hypotension.

\subsection{Model Architectures Under Audit}

We audited two architectures differing fundamentally in their use of historical information (Figure 1A). The two architectures differ in three key dimensions: input conditioning (MedDT includes historical actions; HCT-RL excludes them via Causal Action Shielding), training signal (behavioral cloning vs.\ causal-weighted RL with CQL), and value estimation (none vs.\ CQL-regularized critic).

\subsubsection{Medical Decision Transformer (MedDT)}
MedDT adapts the Decision Transformer\cit{6} to medical sequential decision-making. The model conditions action prediction on the full trajectory including returns-to-go $R_{1:t}$, states $s_{1:t}$, and previous actions $a_{1:t-1}$. At each timestep, returns-to-go, state, and action embeddings are added with timestep encodings and passed through LayerNorm: $e^R_i = \text{LayerNorm}(\text{Linear}_R(R_i) + \text{Embed}_\tau(\tau_i))$, and analogously for $e^s_i$ and $e^a_i$. Tokens are stacked as $[e^R_1, e^s_1, e^a_1, \ldots, e^R_t, e^s_t,\\ e^a_t] \in \mathbb{R}^{3t \times d}$ and processed by a causally masked Transformer encoder ($d=128$, 3 layers, 4 heads). Actions are predicted from state-position outputs via Softplus activation, enforcing non-negative dosages.

The training objective minimizes sequence-level MSE between predicted and clinician actions over all timesteps:
\begin{equation}
\mathcal{L}_{\text{MedDT}} = \mathbb{E}_{(s,a,R)\sim\mathcal{D}} \left[ \frac{1}{t}\sum_{i=1}^{t} \| \hat{a}_i - a_i \|_2^2 \right]
\end{equation}
This architecture is vulnerable to action leakage: by conditioning on $a_{1:t-1}$, the model may shortcut learning by copying previous actions rather than reasoning from physiological states, thereby inheriting the palliative confounder embedded in historical treatment patterns.

\subsubsection{Historical Causal Transformer (HCT-RL)}
HCT-RL addresses action leakage through a multi-component architecture that excludes historical actions from the encoder input (Causal Action Shielding), estimates a behavior policy to construct causal importance weights, and incorporates Conservative Q-Learning (CQL)\cit{5} to penalize over-optimistic value estimates on out-of-distribution actions.

The forward pass employs an action-free state encoder: inputs $s_{1:t}$ are layer-normalized, projected to $\mathbb{R}^{t \times d}$, augmented with positional encodings, and processed through a Transformer encoder to yield latent representation $z_t$ at the final timestep. A multilayer perceptron (MLP) actor maps $z_t$ to $\hat{a}_t$ via Softplus, and an auxiliary MLP decoder predicts the next state $\hat{s}_{t+1}$.

Training proceeds in four coupled steps. First, a propensity network $\hat{\beta}(a_t \mid z_t) = \mathcal{N}(\mu_\beta(z_t), \sigma_\beta^2(z_t))$ estimates the behavior policy by maximizing log-likelihood on observed transitions:
\begin{equation}
\mathcal{L}_{\text{prop}} = -\mathbb{E}_{(s,a)\sim\mathcal{D}}\left[\log \hat{\beta}(a_t \mid z_t)\right]
\end{equation}
Second, causal importance weights are derived from the propensity scores to downweight actions whose frequency in the historical record reflects treatment conventions (including palliative withdrawal) rather than physiological response:
\begin{equation}
w_t = \text{clip}\!\left(\exp\!\left(-\log\hat{\beta}(a_t \mid z_t)\right),\ 0.1,\ 10\right), \quad w_t \leftarrow w_t \,/\, \mathbb{E}[w_t]
\end{equation}
Third, a critic network $Q(z_t, a_t)$ is updated using causal-weighted temporal difference (TD) learning combined with a CQL conservative penalty. With target $y_t = r_t + \gamma(1-d_t)Q_{\text{target}}(z_{t+1}, \hat{a}_{t+1})$:
\begin{equation}
\mathcal{L}_{\text{critic}} = \underbrace{\mathbb{E}\!\left[w_t \cdot (Q(z_t, a_t) - y_t)^2\right]}_{\text{causal-weighted TD}} \;+\; \alpha\, \underbrace{\mathbb{E}\!\left[\log\sum_{\tilde{a}}\exp Q(z_t, \tilde{a}) - Q(z_t, a_t)\right]}_{\mathcal{L}_{\text{CQL}}}
\end{equation}
where $\tilde{a}$ is sampled uniformly from $[0,5]^2$ and from the current actor. The CQL term penalizes $Q$-values of unobserved actions, ensuring conservative policy improvement. Fourth, the actor is updated by jointly minimizing a policy improvement objective, a weighted behavioral cloning (BC) loss emphasizing clinically active dosing decisions ($a_t > 0.01$), and the auxiliary state prediction loss:
\begin{align}
\mathcal{L}_{\text{BC}} &= \mathbb{E}\!\left[\left(1 + \gamma_{\text{act}}\cdot\mathbf{1}[a_t > 0.01]\right) \cdot \|\hat{a}_t - a_t\|_2^2\right] \\[2pt]
\mathcal{L}_{\text{actor}} &= -\mathbb{E}[Q(z_t,\,\hat{a}_t)] \;+\; \beta\,\mathcal{L}_{\text{BC}} \;+\; \lambda\,\mathbb{E}\!\left[\|\hat{s}_{t+1} - s_{t+1}\|_2^2\right]
\end{align}
The causal-weighted TD loss reduces the influence of frequently observed but causally spurious actions (e.g., palliative dose reductions) on the learned value function. The auxiliary state prediction term regularizes the encoder to capture physiological dynamics rather than action-copying shortcuts. Together, these components ensure that policy improvement is rooted in physiological causality rather than distributional frequency.

\subsection{Training and Evaluation}

MedDT was trained with AdamW (learning rate $10^{-4}$, weight decay $10^{-4}$) and a linear warmup scheduler over 1,000 steps. HCT-RL used separate Adam optimizers for the actor (lr $= 10^{-4}$) and critic/propensity (lr $= 3 \times 10^{-4}$). Both models were trained for up to 15 epochs; the checkpoint with lowest validation MSE was retained. Shared architectural parameters: hidden dimension $d=128$, 3 Transformer layers, 4 attention heads, window size $L=6$, batch size 256. Hyperparameter values for HCT-RL: $\alpha=1.0$ (CQL weight), $\beta=2.5$ (BC weight), $\lambda=10.0$ (auxiliary weight), $\gamma_{\text{act}}=20.0$ (active-action emphasis), $\gamma=0.99$ (discount factor).

For standard evaluation, we report MSE on held-out test trajectories and Fitted Q-Evaluation (FQE),\cit{10,11} an off-policy evaluation method that estimates policy value $V^\pi$ by iteratively minimizing the Bellman error:
\begin{equation}
\mathcal{L}_{\text{FQE}}(\phi) = \mathbb{E}_{(s,a,r,s') \sim \mathcal{D}} \left[ \left( Q^\phi(s,a) - r - \gamma Q^{\phi_{\text{target}}}(s', \pi(s')) \right)^2 \right]
\end{equation}
All models were trained across five random seeds (1, 42, 100, 777, 2025). Standard metrics in Table 1 and CCA Audits II--III report results from the seed with lowest validation MSE; Audit I sensitivity in Table 2 reports mean $\pm$ SD across all seeds.

\section{Results}

\subsection{Standard Evaluation Metrics}

Table 1 summarizes performance on MIMIC-III. MedDT achieved substantially lower MSE (0.0098 vs. 0.0742) and higher FQE ($V^\pi = 65.29$ vs. $63.43$), indicating superior behavioral imitation. However, as the following audits reveal, this statistical advantage conceals dangerous misalignment with clinical principles.

\begin{table}[H]
\centering
\caption{Standard Evaluation Metrics for Audited Architectures}
\label{tab:standard_metrics}
\begin{tabular}{lccc}
\toprule
Model & MSE & FQE ($V^\pi$) & Interpretation \\
\midrule
MedDT  & 0.0098 & 65.29 $\pm$ 1.10 & High-fidelity imitation \\
HCT-RL & 0.0742 & 63.43 $\pm$ 0.84 & Causal reconstruction \\
\bottomrule
\end{tabular}
\end{table}

\subsection{Audit I: Spurious Robustness}

Injecting Gaussian noise ($\sigma=0.5$) into creatinine values across all timesteps of the window, MedDT exhibited a sensitivity of 0.058 while HCT-RL achieved 0.021, a 2.8-fold improvement (Figure 2). Relative robustness scores were 96.5\% vs. 100\%. Multi-seed analysis confirmed the stability of this finding (Table 2). MedDT's sensitivity reflects its conditioning on historical action sequences: prior actions encode correlational signals between non-critical laboratory values and treatment patterns, which the model replicates even when those values are perturbed. HCT-RL, reconstructing policy from physiological state trajectories alone, is not susceptible to such spurious correlations.

\begin{figure}[H]
    \centering
    \includegraphics[width=0.68\textwidth]{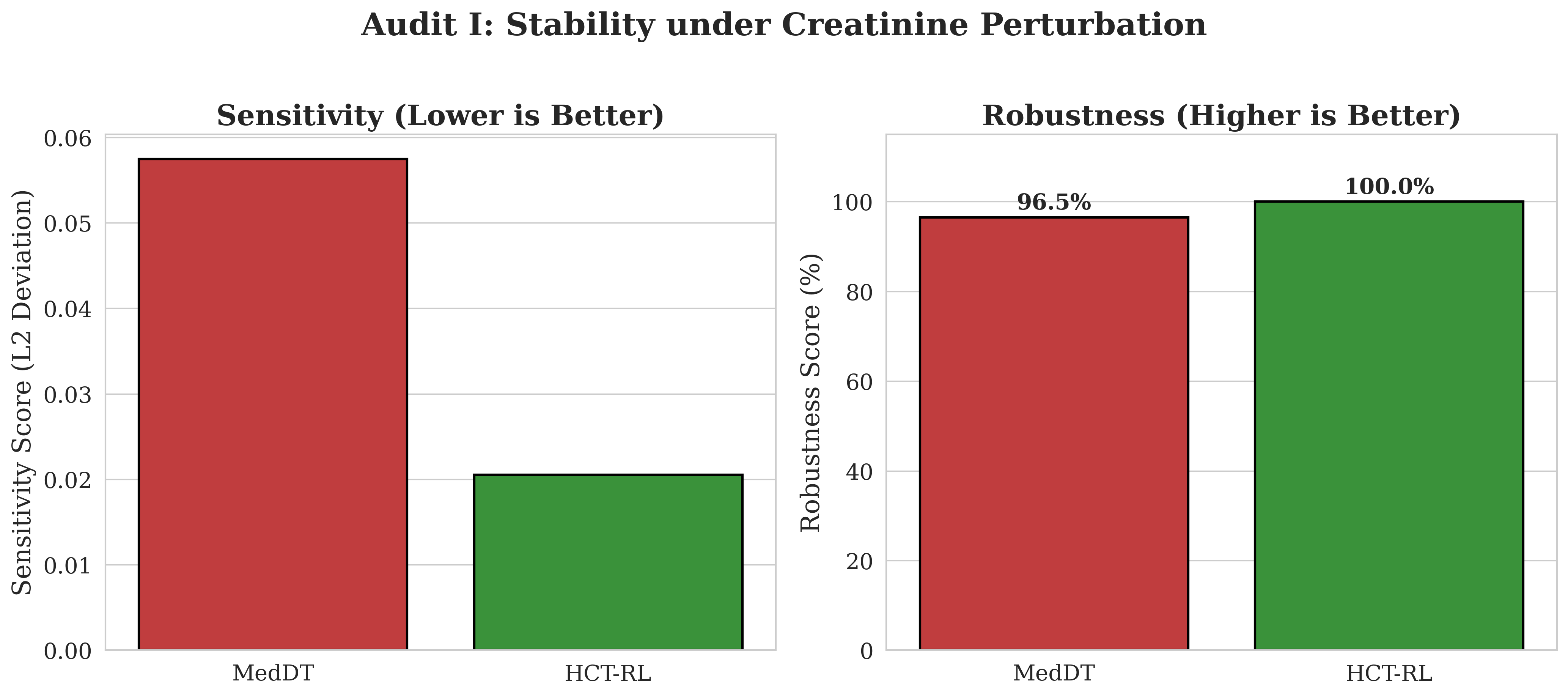}
    \caption{Audit I: Spurious Robustness on MIMIC-III. Left: sensitivity to creatinine perturbation (lower is better). Right: relative robustness scores normalized to the more robust model. HCT-RL demonstrates superior stability under non-critical perturbations.}
    \label{fig:audit1}
\end{figure}

\subsection{Audit II: Causal Trend Alignment}

Counterfactually setting lactate across 15 levels from Z-score $-2$ to $+5$ (applied uniformly across the full observation window), HCT-RL demonstrated a monotonically increasing treatment response, rising approximately 10\% above the baseline level at $\ell_{\min}$, consistent with SSC guidelines recommending lactate-guided dose escalation (Spearman $\rho = +1.00$, reflecting a strictly monotonic response across all 15 lactate levels). MedDT exhibited a paradoxical pattern: treatment intensity fell approximately 15\% below baseline at moderate-to-high lactate levels before a partial recovery at extreme values ($\rho = -0.67$), as shown in Figure 3. This inverse dose-response represents Toxic Mimicry. In the training data, severely elevated lactate frequently co-occurs with transition to palliative care and dose reduction; MedDT, conditioning on historical action sequences, has internalized this confounded association as a treatment rule.

\begin{figure}[H]
    \centering
    \includegraphics[width=0.62\textwidth]{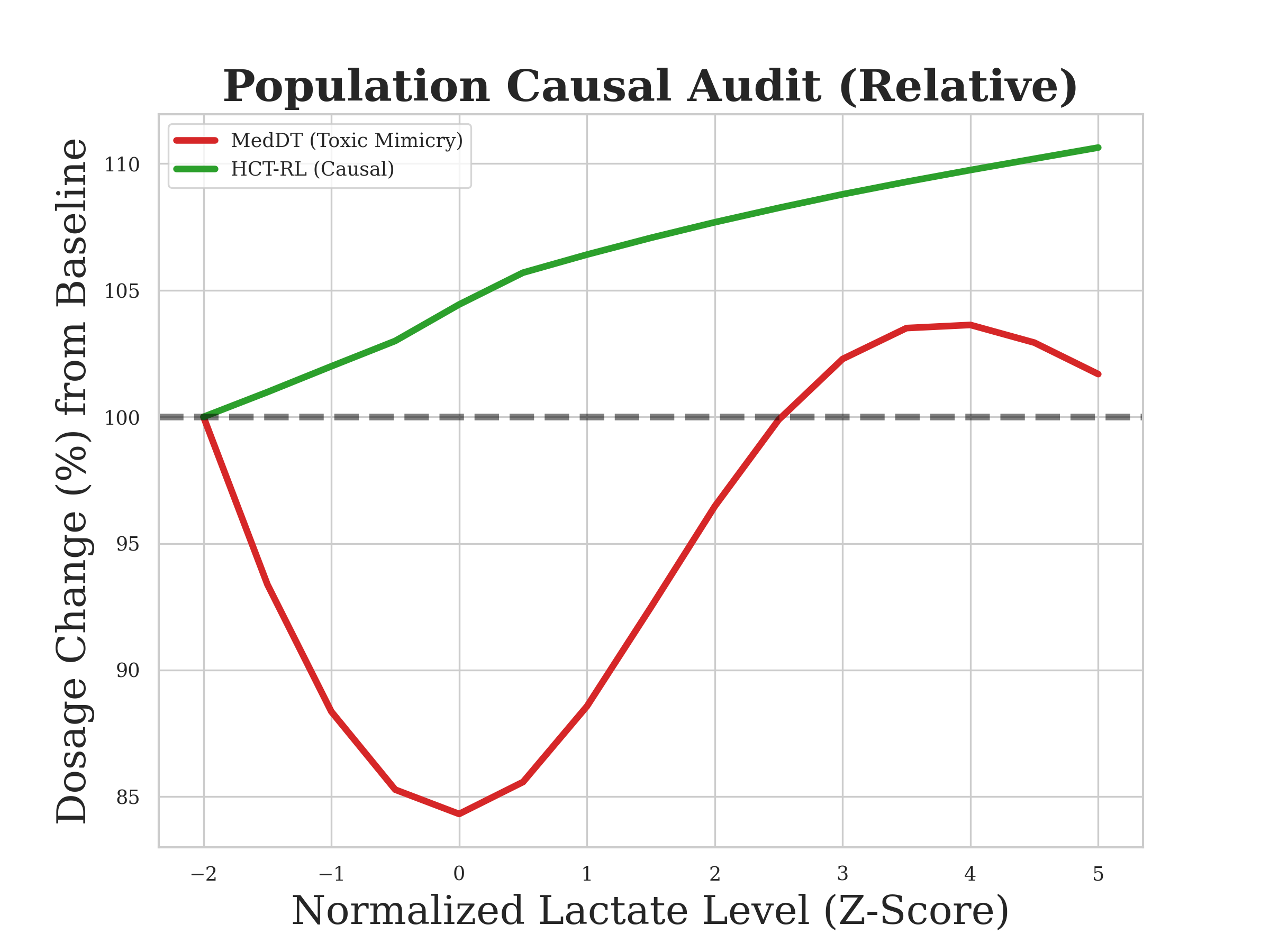}
    \caption{Audit II: Causal Trend Alignment on MIMIC-III. The y-axis shows treatment intensity as percentage change from the baseline response at minimum lactate. HCT-RL shows monotonic dose escalation consistent with SSC guidelines ($\rho = +1.00$). MedDT exhibits paradoxical reduction characteristic of Toxic Mimicry ($\rho = -0.67$).}
    \label{fig:audit2}
\end{figure}

\subsection{Audit III: Contextual Scissor Probe}

Sweeping MAP from Z=$+1$ to Z=$-2$, HCT-RL exhibited pronounced Scissor Divergence: its vasopressor response steepened markedly in Scenario B relative to Scenario A, as shown in Figure 4. This steepening reflects appropriate integration of multi-organ stress signals; the model correctly recognizes that identical MAP levels carry higher urgency when accompanied by tachycardia and hyperlactatemia. MedDT showed minimal contextual discrimination, producing near-identical vasopressor slopes across both scenarios ($\Delta_{\text{scissor}} \approx 0$), failing to differentiate isolated hypotension from distributive shock with multi-organ compromise.

\begin{figure}[H]
    \centering
    \includegraphics[width=0.66\textwidth]{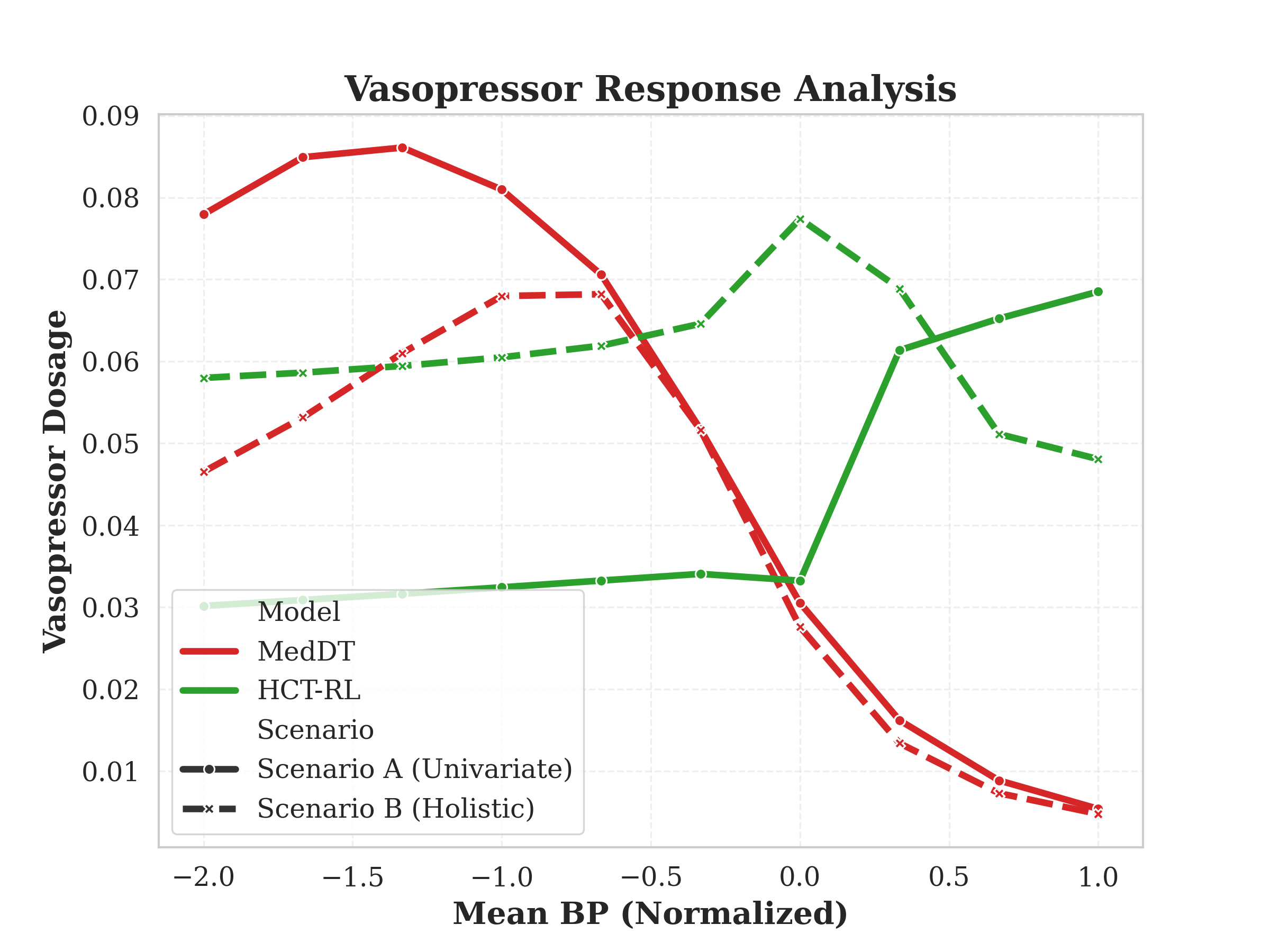}
    \caption{Audit III: Contextual Scissor Probe on MIMIC-III. The upper panel shows mean vasopressor dosage vs. MAP for both scenarios. The lower panel shows per-step Scissor Divergence (Holistic $-$ Univariate). HCT-RL exhibits consistently positive divergence, scaling urgency under multi-organ stress. MedDT fails to discriminate clinical contexts.}
    \label{fig:audit3}
\end{figure}

\subsection{External Validation on MIMIC-IV}

To assess generalizability, we evaluated on MIMIC-IV (v3.1)\cit{22,23} without retraining, testing whether observed patterns reflect architectural properties rather than dataset-specific artifacts.

\subsubsection{Audit I on MIMIC-IV}
The performance gap was amplified under distribution shift: MedDT sensitivity increased to 0.10 while HCT-RL achieved 0.01, a 10-fold improvement (Figure 5). Relative robustness scores: 91.9\% vs. 100\%. Multi-seed analysis confirmed this pattern with even wider separation (Table 2). As distributional distance from the training set increases, MedDT's reliance on historical action correlations becomes more hazardous, whereas HCT-RL's causal grounding in physiological state trajectories provides robust generalization. Recent work has independently documented similar transportability challenges when applying ICU-trained RL models to new clinical settings.\cit{26}

\begin{figure}[H]
    \centering
    \includegraphics[width=0.68\textwidth]{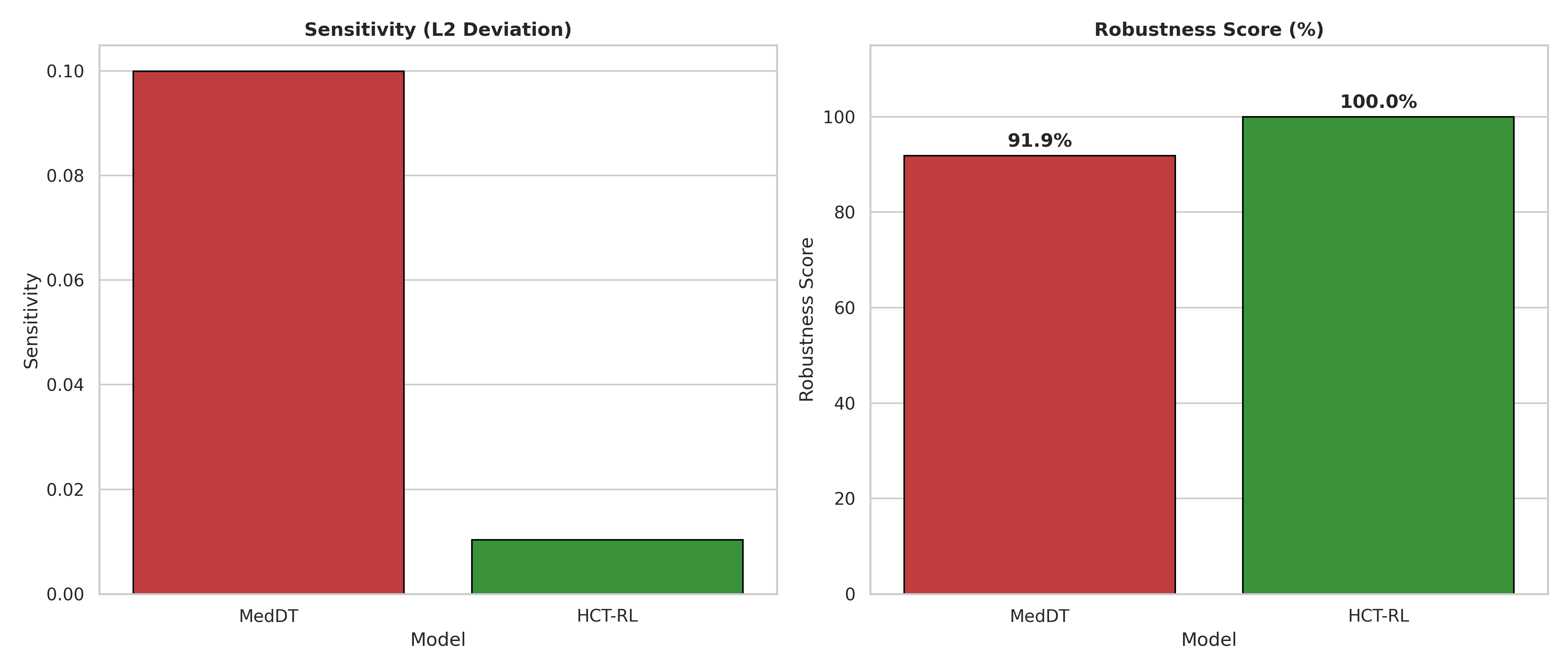}
    \caption{Audit I on MIMIC-IV: Performance gap amplified under distribution shift. HCT-RL demonstrates 10-fold lower sensitivity than MedDT, confirming architectural robustness rather than dataset-specific adaptation.}
    \label{fig:audit1_mimic4}
\end{figure}

\subsubsection{Audit III on MIMIC-IV}
HCT-RL maintained positive Scissor Divergence (0.04--0.12) across the MAP sweep on MIMIC-IV (Figure 6). MedDT exhibited negative divergence ($-0.12$ to 0): under multi-organ stress, it paradoxically recommended lower vasopressor doses than in the univariate condition, inverting the clinically required response. This pattern represents an especially dangerous failure mode under distribution shift, where learned confounders from the original training cohort are applied to a new patient population.

\begin{figure}[H]
    \centering
    \includegraphics[width=0.50\textwidth]{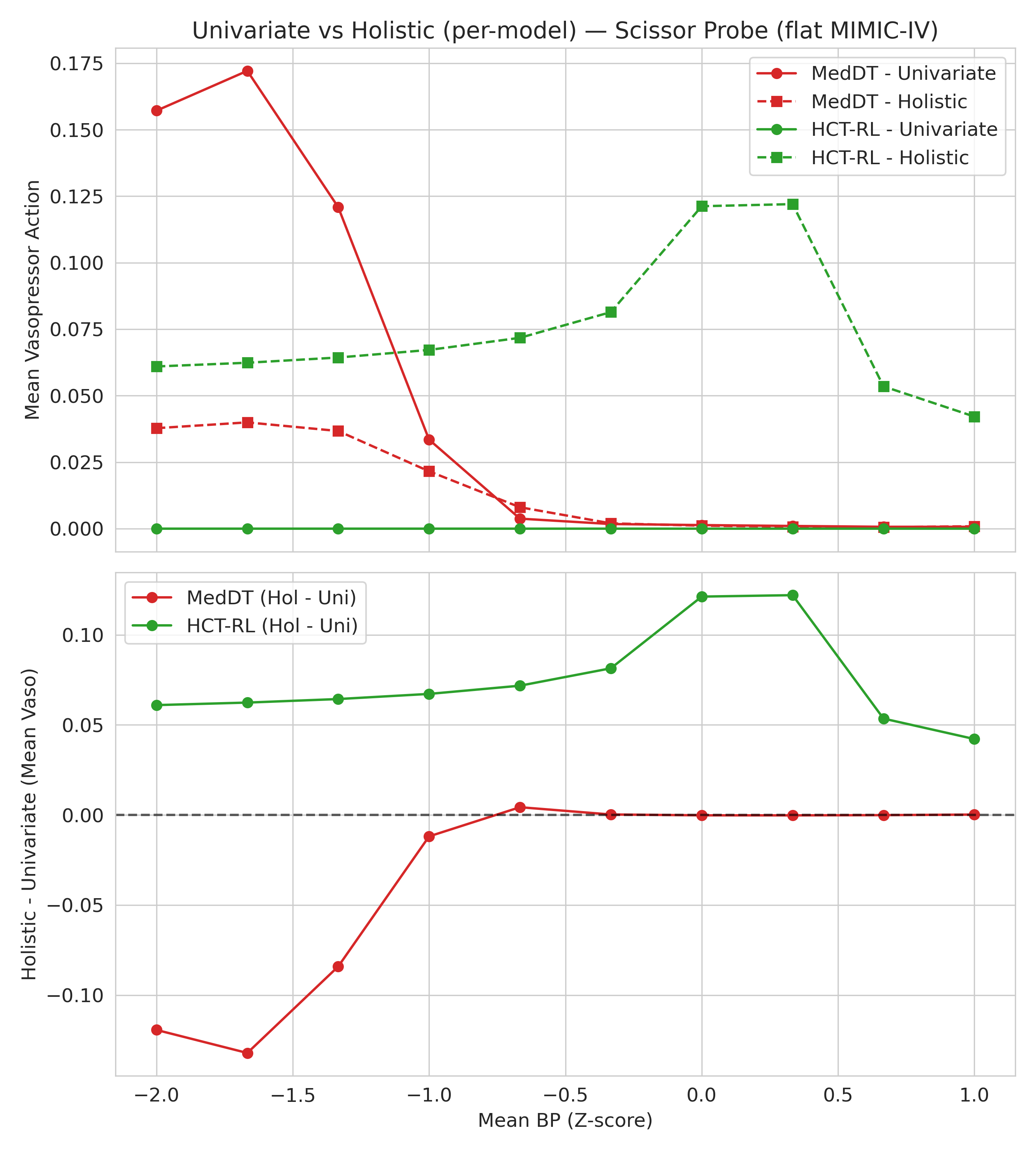}
    \caption{Audit III on MIMIC-IV: HCT-RL maintains positive Scissor Divergence throughout the MAP range. MedDT exhibits negative divergence, paradoxically reducing vasopressor support under multi-organ stress.}
    \label{fig:audit3_mimic4}
\end{figure}

\subsection{Summary of CCA Findings}

Table 2 summarizes results across both datasets. MedDT achieved superior standard metrics but failed all CCA audits. The external validation confirms that Toxic Mimicry is a systematic failure mode arising from architectural dependence on historical actions, not a dataset-specific artifact.

\begin{table}[H]
\centering
\footnotesize
\caption{Summary of Counterfactual Clinical Audit Results. Audit I sensitivity values are mean $\pm$ SD across five random seeds; figures present results from a representative seed.}
\label{tab:cca_summary}
\begin{tabular}{llcccc}
\toprule
 & & \multicolumn{2}{c}{MIMIC-III} & \multicolumn{2}{c}{MIMIC-IV} \\
\cmidrule(lr){3-4} \cmidrule(lr){5-6}
Audit & Metric & MedDT & HCT-RL & MedDT & HCT-RL \\
\midrule
I: Spurious      & Sensitivity & 0.060$\pm$0.006 & 0.032$\pm$0.008 & 0.062$\pm$0.028 & 0.003$\pm$0.003 \\
Robustness       & Relative Score & 97.4\% & 100\% & 94.4\% & 100\% \\
\midrule
II: Causal       & Spearman $\rho$ & $-0.67$ & $+1.00$ & -- & -- \\
Trend            & Toxic Mimicry & Yes & No & -- & -- \\
\midrule
III: Scissor     & Divergence $\Delta$ & $\approx 0$ & Positive & Negative & Positive \\
Probe            & Clinical Logic & Fail & Pass & Fail & Pass \\
\bottomrule
\end{tabular}
\end{table}

\section{Discussion and Conclusion}

\subsection{Discussion}

This study reveals a systematic decoupling between standard evaluation metrics and clinical safety in medical offline RL. MedDT achieved superior MSE and FQE yet failed all CCA audits. This paradox exposes a core limitation: metrics rewarding faithful replication of historical behavior cannot distinguish clinically appropriate patterns from harmful confounders. The palliative confounder causes high-capacity sequence models to interpret treatment withdrawal as valid therapeutic responses, producing Toxic Mimicry where policies recommend de-escalation precisely when guidelines mandate intensification. Moreover, because end-of-life care decisions are known to vary across racial, ethnic, and socioeconomic groups,\cit{17,18,19} the palliative confounder is unlikely to be uniformly distributed. Toxic Mimicry may therefore disproportionately affect patient subpopulations already subject to disparities in ICU care; if palliative transitions are more frequent in underserved populations, the resulting systematic undertreatment would compound existing disparities in sepsis outcomes.

Toxic Mimicry shares conceptual roots with causal confusion in imitation learning, where policies learn spurious correlates from expert demonstrations that fail under distributional shift.\cit{27} In the broader offline RL literature, this belongs to the class of reward misspecification under unobserved confounding. Our contribution is to expose this failure mode in a clinical domain where the confounding mechanism is identifiable and guideline-mandated dose-response directions enable falsifiable audits rather than relying on post-hoc reward analysis.

The architectural contrast between MedDT and HCT-RL reveals the underlying mechanism. MedDT's conditioning on $a_{1:t-1}$ allows it to absorb the full distribution of historical treatment patterns, including palliative care trajectories. HCT-RL's Causal Action Shielding forces policy reconstruction from physiological state trajectories alone. The causal importance weighting further suppresses the influence of high-frequency palliative actions on Q-function learning, while the CQL term prevents overestimation of conservative dose-reduction strategies. That HCT-RL maintains comparable FQE (63.43 vs. 65.29) despite substantially higher MSE, while passing all CCA audits, demonstrates that causal consistency and standard performance are not inherently in conflict; rather, standard metrics are insufficient to detect clinically serious failures.

The CCA framework shifts evaluation from behavioral imitation to causal consistency, asking whether dose-response relationships align with physiological principles encoded in clinical guidelines. The external validation on MIMIC-IV shows that Toxic Mimicry amplifies under distribution shift, confirming the need for causal auditing as a prerequisite rather than an optional supplement to standard benchmarks.

Beyond sepsis, CCA applies to any ICU decision domain where clinical guidelines establish a causal dose-response structure. Mechanical ventilation in acute respiratory distress syndrome (ARDS) provides a direct structural analogue: the ARDSNet trial established that low tidal volume ventilation (6 mL/kg) reduces mortality in patients with acute lung injury (PaO$_2$/FiO$_2$ [P/F] $\leq$300), such that a policy recommending tidal volumes above this protective threshold as oxygenation worsens would constitute an analogous failure mode to Audit II.\cit{24} Similar structures exist in insulin infusion protocols, sedation depth management, and antibiotic de-escalation, where treatment decisions are governed by physiologically motivated dose-response principles.\cit{2,8} CCA can be instantiated wherever guidelines provide falsifiable causal predictions, requiring only a domain-specific Audit II constructed from the relevant clinical criteria.

A key limitation of the current work is that CCA audits behavioral safety, specifically whether a policy's dose-response logic is causally consistent with clinical guidelines, but cannot directly certify outcome improvement. The gap between causal consistency and patient outcomes is well-established even within guideline-adherent therapies: the VASST trial demonstrated that in patients with septic shock who all received vasopressor support meeting SSC hemodynamic targets, the choice of vasopressor agent still produced heterogeneous individual outcomes, showing that guideline compliance does not eliminate individual-level treatment variability.\cit{25} Outside the ICU, where reward signals are sparser and confounding more pervasive, this gap widens further.\cit{8,9} CCA is therefore best understood as a necessary safety screen rather than a sufficient criterion: policies that fail causal audits exhibit clinically inconsistent behavior that is incompatible with safe deployment, but those that pass still require prospective clinical validation before deployment.\cit{1}

Limitations include: CCA audits anchored in SSC guidelines may not capture all dimensions of optimal management; counterfactual interventions assume independent manipulation of biomarkers; and evaluation is limited to two architectures. Both MIMIC datasets reflect a single academic medical center in Boston, whose institutional patterns and demographic characteristics may differ from other health systems; model behavior under the palliative confounder may differ in populations with different racial, ethnic, and socioeconomic compositions, particularly given documented disparities in end-of-life care decision-making across these groups. The importance weight clipping bounds (0.1, 10) follow standard offline RL practice; sensitivity to propensity model misspecification and alternative clipping ranges was not systematically evaluated. Cross-seed variability of Audit II and III metrics, which involve nonlinear counterfactual sweeps, warrants further investigation.

\subsection{Conclusion}

This study identifies Toxic Mimicry as a critical failure mode in medical offline RL and introduces the CCA framework to expose safety hazards that conventional benchmarks cannot detect. HCT-RL demonstrates that integrating Causal Action Shielding, propensity-based importance weighting, and CQL effectively suppresses palliative confounders, maintaining physiological consistency where standard Decision Transformers fail. This work calls for a shift in evaluating AI-driven clinical decision support: predictive accuracy is insufficient for safety-critical applications; counterfactual audits anchored in domain knowledge must become standard. The ultimate goal of medical RL is not to replicate historical decisions but to discover strategies that improve patient outcomes, requiring a transition from imitation toward causal reasoning.

\begin{center}
\textbf{References}
\end{center}
\begin{enumerate}[label={\arabic*.}, leftmargin=*, noitemsep, topsep=0pt]

    \item Gottesman O, Johansson F, Komorowski M, et al. Guidelines for reinforcement learning in healthcare. \textit{Nat Med}. 2019;25(1):16-18.
    \item Liu S, See KC, Ngiam KY, Celi LA, Sun X, Feng M. Reinforcement learning for clinical decision support in critical care: comprehensive review. \textit{J Med Internet Res}. 2020;22(7):e18477.
    \item Komorowski M, Celi LA, Badawi O, Gordon AC, Faisal AA. The artificial intelligence clinician learns optimal treatment strategies for sepsis in intensive care. \textit{Nat Med}. 2018;24(11):1716-1720.
    \item Raghu A, Komorowski M, Ahmed I, Celi LA, Szolovits P, Ghassemi M. Deep reinforcement learning for sepsis treatment [Preprint]. arXiv:1711.09602. 2017.
    \item Kumar A, Zhou A, Tucker G, Levine S. Conservative Q-learning for offline reinforcement learning. \textit{NeurIPS}. 2020;33:1179-1191.
    \item Chen L, Lu K, Rajeswaran A, et al. Decision transformer: reinforcement learning via sequence modeling. \textit{NeurIPS}. 2021;34:15084-15097.
    \item Roggeveen LF, El Hassouni A, de Grooth HJ, Girbes ARJ, Hoogendoorn M, Elbers PWG. Reinforcement learning for intensive care medicine: actionable clinical insights from novel approaches to reward shaping and off-policy model evaluation. \textit{Intensive Care Med Exp}. 2024;12(1):32.
    \item Yu C, Liu J, Nemati S, Yin G. Reinforcement learning in healthcare: a survey. \textit{ACM Comput Surv}. 2021;55(1):1-36.
    \item Gottesman O, Johansson F, Meier J, et al. Evaluating reinforcement learning algorithms in observational health settings [Preprint]. arXiv:1805.12298. 2018.
    \item Le H, Voloshin C, Yue Y. Batch policy learning under constraints. \textit{ICML}. 2019:3703-3712.
    \item Tang S, Wiens J. Model selection for offline reinforcement learning: practical considerations for healthcare settings. \textit{PMLR}. 2021;149:2-35.
    \item Richens JG, Lee CM, Johri S. Improving the accuracy of medical diagnosis with causal machine learning. \textit{Nat Commun}. 2020;11(1):3923.
    \item Prosperi M, Guo Y, Sperrin M, et al. Causal inference and counterfactual prediction in machine learning for actionable healthcare. \textit{Nat Mach Intell}. 2020;2(7):369-375.
    \item Sanchez P, Voisey JP, Xia T, Watson HI, O'Neil AQ, Tsaftaris SA. Causal machine learning for healthcare and precision medicine. \textit{R Soc Open Sci}. 2022;9(8):220638.
    \item Singer M, Deutschman CS, Seymour CW, et al. The third international consensus definitions for sepsis and septic shock (Sepsis-3). \textit{JAMA}. 2016;315(8):801-810.
    \item Evans L, Rhodes A, Alhazzani W, et al. Surviving sepsis campaign: international guidelines for management of sepsis and septic shock 2021. \textit{Intensive Care Med}. 2021;47(11):1181-1247.
    \item Prendergast TJ, Claessens MT, Luce JM. A national survey of end-of-life care for critically ill patients. \textit{Am J Respir Crit Care Med}. 1998;158(4):1163-1167.
    \item Sprung CL, Cohen SL, Sjokvist P, et al. End-of-life practices in European intensive care units: the Ethicus study. \textit{JAMA}. 2003;290(6):790-797.
    \item Mark NM, Rayner SG, Lee NJ, Curtis JR. Global variability in withholding and withdrawal of life-sustaining treatment in the intensive care unit: a systematic review. \textit{Intensive Care Med}. 2015;41(9):1572-1585.
    \item Johnson A, Pollard T, Mark R. MIMIC-III clinical database (version 1.4). PhysioNet. 2016. doi:10.13026/C2XW26
    \item Johnson AEW, Pollard TJ, Shen L, et al. MIMIC-III, a freely accessible critical care database. \textit{Sci Data}. 2016;3:160035.
    \item Johnson A, Bulgarelli L, Pollard T, et al. MIMIC-IV (version 3.1). PhysioNet. 2024. doi:10.13026/kpb9-mt58
    \item Johnson AEW, Bulgarelli L, Shen L, et al. MIMIC-IV, a freely accessible electronic health record dataset. \textit{Sci Data}. 2023;10:1.
    \item Acute Respiratory Distress Syndrome Network. Ventilation with lower tidal volumes as compared with traditional tidal volumes for acute lung injury and the acute respiratory distress syndrome. \textit{N Engl J Med}. 2000;342(18):1301-1308.
    \item Russell JA, Walley KR, Singer J, et al. Vasopressin versus norepinephrine infusion in patients with septic shock. \textit{N Engl J Med}. 2008;358(9):877-887.
    \item Nauka PC, Kennedy JN, Brant EB, et al. Challenges with reinforcement learning model transportability for sepsis treatment in emergency care. \textit{npj Digit Med}. 2025;8(1):91.
    \item de Haan P, Jayaraman D, Levine S. Causal confusion in imitation learning. \textit{NeurIPS}. 2019;32.
    
\end{enumerate}

\end{document}